\documentclass[conference]{IEEEtran}
\IEEEoverridecommandlockouts
\usepackage{cite}
\usepackage{amsmath,amssymb,amsfonts}
\usepackage{algorithmic}
\usepackage{graphicx}
\usepackage{textcomp}
\usepackage{xcolor}
\def\BibTeX{{\rm B\kern-.05em{\sc i\kern-.025em b}\kern-.08em
    T\kern-.1667em\lower.7ex\hbox{E}\kern-.125emX}}
    
\usepackage{bm}
\usepackage[linesnumbered,ruled,vlined]{algorithm2e}
\usepackage{multirow}
\usepackage{subfigure}
\usepackage{bbm}
\usepackage{booktabs}
\usepackage{filecontents}
\usepackage{subfigure} %子图
\usepackage{fancyhdr}
\usepackage{flushend}

\usepackage{threeparttable}
\usepackage{enumitem} %\for {itemize}[leftmargin=*]

\usepackage{marvosym}

\newtheorem{definition}{\bf \emph{Definition}}
\newtheorem{theorem}{\bf \emph{Theorem}}

\begin{document}

\title{Prism: Mining Task-aware Domains in Non-\emph{i.i.d.} IMU Data for Flexible User Perception}

\author{Yunzhe Li$^{1*}$, Facheng Hu$^{1*}$\thanks{$^*$ Yunzhe Li and Facheng Hu contributed equally to this paper.}, Hongzi Zhu$^{1\S}$\thanks{$^\S$ Hongzi Zhu is the corresponding author of this paper.}, Quan Liu$^{1,2}$,\\
Xiaoke Zhao$^3$, Jiangang Shen$^1$, Shan Chang$^4$, Minyi Guo$^1$\\
$^1$Shanghai Jiao Tong University, $^2$Nanyang Technological University,\\
$^3$Ant Group, $^4$Donghua University\\
\{yunzhe.li, facheng\_hu, hongzi\}@sjtu.edu.cn}
\maketitle

% \renewcommand{\shortauthors}{Yunzhe Li, Facheng Hu and et al.}

\begin{abstract}
A wide range of user perception applications leverage inertial measurement unit (IMU) data for online prediction.
However, restricted by the non-\emph{i.i.d.} nature of IMU data collected from mobile devices, most systems work well only in a controlled setting (\emph{e.g.,} for a specific user in particular postures), limiting application scenarios. To achieve uncontrolled online prediction on mobile devices, referred to as the flexible user perception (FUP) problem, is attractive but hard.
In this paper, we propose a novel scheme, called \emph{Prism}, which can obtain high FUP accuracy on mobile devices.
The core of Prism is to discover \emph{task-aware} domains embedded in IMU dataset, and to train a domain-aware model on each identified domain. To this end, we design an expectation-maximization (EM) algorithm to estimate latent domains with respect to the specific downstream perception task. Finally, the best-fit model can be automatically selected for use by comparing the test sample and all identified domains in the feature space.
We implement Prism on various mobile devices and conduct extensive experiments. Results demonstrate that Prism can achieve the best FUP performance with a low latency.
\end{abstract}





\section{Introduction}
Recent years have witnessed the soaring development of appealing user perception applications on smart mobile devices, such as user authentication ~\cite{shi2021face, xu2020touchpass, zhu2017shakein}, activity recognition~\cite{chen2021magx, ouyang2022cosmo, ouyang2021clusterfl}, and health monitoring~\cite{cao2021itracku, narayana2021sos}, where machine learning models trained on collected inertial measurement unit (IMU) data are leveraged for online prediction. 
In general, the successes of these user perception applications rely on the superior performance of deep neural networks (DNNs), trained on independent and identically distributed (\emph{i.i.d.}) datasets \cite{pan2009survey}, largely limiting the application scenarios in a controlled setting (\emph{e.g.,} for a specific user in particular postures).
However, datasets flexibly collected from mobile devices are often the case \emph{non-\emph{i.i.d.}} because of different device types and usage habits \cite{xu2023practically}. \emph{Can we achieve flexible user perception (FUP) by training DNNs on IMU data flexibly collected from different types of devices and distinct users without requiring how they operate their devices?}
%For instance, a phone can recognize a legitimate user based on IMU data, no matter whether the phone is held in either hand while walking or standing still.

An attractive scheme to the FUP problem is demanding due to the following reasons. First, it should be able to deal with IMU data collected from multiple non-\emph{i.i.d.} sources (\emph{e.g.}, a device held in different postures or a device of a different brand carried by a different user). Such a dataset contains multiple \emph{hidden} distributions (or domains) \cite{li2021cross, zhao2011cross}, making it hard to train an effective DNN. Second, it should achieve satisfactory prediction accuracy with few constraints on how devices are operated. Third, such a scheme should be lightweight and can be easily deployed to a wide variety of mobile devices with limited computational capacity.

In the literature, much effort has been made to improve the accuracy of FUP on mobile devices. One main direction aims to develop one single prediction model that can generalize on all potential domains via domain generalization methods \cite{lu2023out, ganin2016domain}, meta-learning \cite{bi2022csear} and pre-training \cite{xu2021limu, tripathi2022imair, bhalla2021imu2doppler}. However, their performance improvements are marginal because the essential non-\emph{i.i.d.} issue still exists \cite{Qian2022, haresamudram2022assessing}. 
Another direction is to divide training data into subsets and train an individual model on each subset, respectively. One or a few of those models best fitting the current scenario are selected for online prediction. One class of methods divides training dataset manually based on some prior knowledge  \cite{zamir2018taskonomy, fang2019teamnet, chen2022intent} (\emph{e.g.}, user intention in recommendation system \cite{chen2022intent} or image quality in computer vision \cite{zamir2018taskonomy}) or associated attributes of data samples (\emph{i.e.}, metadata) such as where a device is carried or the user ID \cite{niu2020billion, li2021hermes}. However, to obtain meaningful metadata is of intensive manpower and how to select effective metadata to use is not straightforward. Another class of methods clusters similar data samples either in raw data space \cite{zhang2016self, zheng2019metadata, zheng2022life, li2024anole} or in some high-dimensional feature space \cite{chen2022intent, caron2018deep, li2020prototypical}. However, the derived subsets may not match the latent distributions with respect to a particular user perception task. As a result, to the best of our knowledge, there is no existing scheme that successfully addresses the FUP problem.

\begin{figure}
	\centering
	\includegraphics[width=\linewidth]{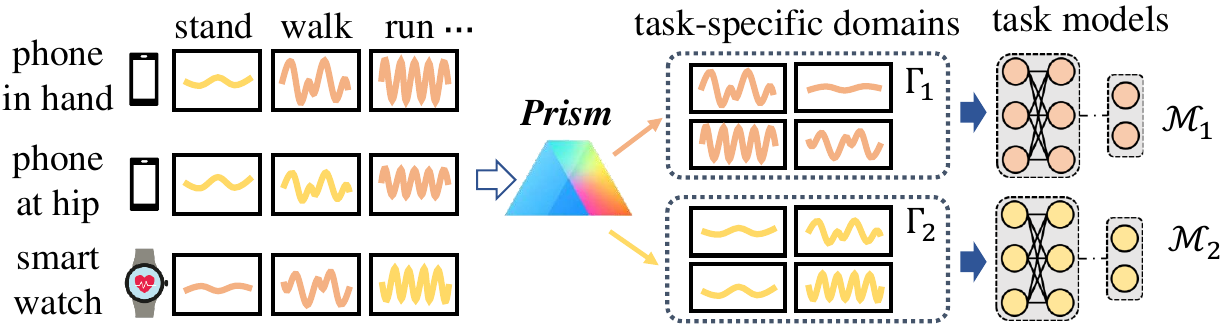}
	\caption{Illustration of IMU data partition using Prism, where non-\emph{i.i.d.} samples are divided into task-specific domains, denoted with different colors, rather than prior-defined subsets according to device positions or types.}
	\label{fig:omi}
\end{figure}

In this paper, we propose an effective data partition scheme, called \emph{Prism}, which measures the inconsistence extent of a non-\emph{i.i.d.} dataset and wisely divides the dataset into domains friendly to a downstream user perception task.
Given a non-\emph{i.i.d.} dataset, we have an insight that different tasks (or corresponding DNN models) may have distinct domain partitions. The core idea of Prism, therefore, is to \emph{automatically find a feature space where similar samples form \emph{i.i.d.} domains for a particular downstream task}. Then, individual models can be well-trained on each \emph{task-specific} domain. As illustrated in Figure \ref{fig:omi}, in Prism, non-\emph{i.i.d.} data samples are first divided into task-specific domains before task models can be trained. When conducting uncontrolled online prediction, a best-fit trained model can be selected for use by comparing the test sample and all identified task-specific domains in the feature space.

The Prism design faces two main challenges.
First, it's hard to tell whether a given dataset is non-\emph{i.i.d.} (\emph{i.e.}, containing multiple distributions) without any prior information. To deal with this challenge, we propose a \emph{non-\emph{i.i.d.} degree} of a dataset (NID) as the quantitative measure of non-\emph{i.i.d.} NID is calculated by testing the prediction inconsistency within the dataset. Specifically, we first divide the dataset into two parts and calculate a \emph{non-\emph{i.i.d.} index} (NI) between the divided two parts. The data samples between the two parts are then alternated to obtain a traversal of the dataset and obtain multiple NIs. Finally, NID is defined as the average of the multiple NIs during the traversal of the dataset.

Second, task-specific domains are latent, which means there is no obvious clue to estimate them in a non-\emph{i.i.d.} dataset. Indeed, to find the optimal task-specific domains is NP-hard. 
To tackle this challenge, we design a neat Expectation-Maximization (EM) algorithm to iteratively train an encoder, with which data samples can be converted into embeddings in the feature space. Moreover, clusters of similar embeddings can be used to correspondingly train a set of downstream task models with the best performance. Specifically, each iteration consists of an estimation-step (E-step) and a maximization-step (M-step). In the E-step, $k$-means is adopted to group similar embeddings obtained by the encoder from the previous iteration into distinct clusters. In the M-step, an individual task model is first trained on each cluster. Then, we assess both the quality of the derived clusters and that of obtained task models via trial tests, both of which are utilized to design a joint loss to optimize the parameters of the encoder. In this way, after the EM algorithm converges, we can obtain a superb estimation of latent task-specific domains.

%\revised{Third, for uncontrolled online prediction, it is inevitable that an arbitrary test sample may be misclassified into a wrong domain, missing the best-fit task model for prediction. This misclassification can have significant consequences, given that each task model is trained solely within its own domain (\emph{i.e.}, native domain) and lacks the ability to generalize to other domains (\emph{i.e.}, non-native domain).
	%To address this challenge, we propose a multi-task learning approach that leverages knowledge transfer between domains during task model training. This enables each task model to possess a basic capacity to handle test samples from domains other than its own, while also being optimally equipped to handle samples within its native domain.
	%To this end, we employ a shared backbone (for feature extraction) for all task models, with each task model possessing its own unique classifier for its respective domain. The samples are first processed through the shared global backbone and are subsequently directed to the classifiers of their specific domains.
	%As a result, Prism facilitates effective knowledge transfer between diverse domains, thereby enabling reliable prediction accuracy for both native and non-native test samples.}

We implement Prism on 6 typical mobile devices with different CPU/GPU configurations.
We consider activity recognition (AR) and user authentication (UA) as typical user perception tasks, and conduct extensive experiments on non-\emph{i.i.d.} public IMU datasets, \emph{i.e.}, UCI \cite{reyes2016transition}, HHAR \cite{stisen2015smart}, and Motion \cite{malekzadeh2019mobile}. We also construct a more non-\emph{i.i.d.} large-scale dataset based on the extensive SHL dataset \cite{gjoreski2017versatile} to show the FUP performance of Prism on more complicated settings. Experiment results show that Prism can effectively estimate the latent task-specific domains, achieving reliable and state-of-the-art (SOTA) prediction accuracy for flexible user perception applications. 
Results demonstrate that Prism can achieve reliable FUP prediction and outperform a universal deep model in terms of F1 score on datasets with high NID, achieving an improvement of up to 16.79\%.
Prism is lightweight and can be easily deployed on most mobile devices, with a latency less than 60 ms even on a low-end smartphone.

We highlight the main contributions made in this paper as follows: 
\begin{itemize}
    \item A non-\emph{i.i.d.} degree of a dataset is delicately designed to quantify the non-\emph{i.i.d.} level of a complex dataset;
    \item The NP-hardness of automatically finding latent domains is analyzed, and Prism, a joint method for task-specific domain partition and corresponding task model training based on an EM algorithm, is proposed;
    \item Prism is implemented on various types of mobile devices and evaluated on multiple public IMU datasets. Results demonstrate the efficacy of Prism's design. 
\end{itemize}
% 1) A non-\emph{i.i.d.} degree of a dataset is delicately designed to quantify the non-\emph{i.i.d.} level of a complex dataset; 2) The NP-hardness of automatically finding latent domains is analyzed, and Prism, a joint method for task-specific domain partition and corresponding task model training based on EM algorithm, is proposed; 3) Prism is implemented on various types of mobile devices and evaluated on multiple public IMU datasets and a real-world large-scale IMU dataset. Results demonstrate the efficacy of Prism's design. 

\section{Problem Definition}
% \subsection{System Model}
% We consider three types of entities in the system:
% \begin{itemize}[leftmargin=*]
% 	\item \textbf{Cloud server}: has sufficient computational power and storage for offline model training.
% 	\item \textbf{Mobile devices}: possess constrained computational power and limited memory while carrying out online user perception. These devices are outfitted with low-end IMU sensors configured at different sampling rates and battery-powered, demanding lightweight algorithms. Furthermore, these devices are connected to the cloud server via wireless communication.
% 	\item \textbf{Users}: do not need to cooperate with a specific perception application such as AR and UA and can operate their devices in an uncontrolled way.
% \end{itemize}

% \subsection{Problem Formulation}

Given a dataset of IMU samples collected from different users, denoted as $D$, there exists a data partition scheme $\mathcal{P}$, which separates $D$ into $n$ subsets, denoted as $\{\Gamma_1, \Gamma_2, \cdots, \Gamma_n\}$. For each $\Gamma_i$, for $i\in[1,n]$, a task model $\mathcal{M}_i$ can be trained on the training set of $\Gamma_i$, denoted as $\Gamma_i^{\text{trn}}$. The FUP problem can be defined as follows:
\begin{definition}
The FUP problem is to find an optimal data partition scheme, denoted as $\mathcal{P}^*$, so that the prediction errors of testing each obtained task model $\mathcal{M}_i$ on the corresponding testing set of $\Gamma_i$, denoted as $\Gamma_i^{\text{tst}}$,  \emph{i.e.}, $\sum_{i=1}^n{\mathcal{E}(\mathcal{M}_i, \Gamma_i^{\text{tst}})}$, is minimized, where $\mathcal{E}(\mathcal{M}_i,\Gamma_i^{\text{tst}})$ denotes the prediction error of testing $\mathcal{M}_i$ on $\Gamma_i^{\text{tst}}$.
\end{definition}

The FUP problem is hard when dataset $D$ contains non-\emph{i.i.d.} distributions (\emph{e.g.}, $D$ is collected from multiple subjects with different devices) as data distributions captured by DNN models are latent. We have the following theorem:

\begin{theorem}
\label{the:npc}
The FUP problem is NP-hard.
\end{theorem}

\begin{IEEEproof}
The FUP problem can be reduced from the weighted set cover problem \cite{cygan09}, a classic NP problem. Specifically, let $U$ denote a set of $N$ elements, \emph{i.e.}, $U = \{u_1, u_2, \cdots, u_N\}$ and $C(S_i)$ denote the cost of $S_i$ in the power set of $U$, \emph{i.e.}, $\wp(U) = \{S_1, S_2, \cdots, S_{2^N}\}$, for $i \in [1, 2^N]$.
Given $U$ and $\wp(U)$, the objective of the weighted set cover problem is to infer a subset of $\wp(U)$, denoted as $\mathcal{K}$, where  $\bigcup_{S_i \in \mathcal{K}} S_i = U$ so that the sum of the cost of $U_i$ for $S_i \in \mathcal{K}$, \emph{i.e.}, $\sum_{S_i \in \mathcal{K}}C(S_i)$, is minimized. We regard data samples $\{x_1, x_2, \cdots, x_N\}$ in dataset $D$ as the elements $\{u_1, u_2, \cdots, u_N\}$ in $U$. Similarly, the power set of $D$, $\wp(D) = \{\Gamma_1, \Gamma_2, ..., \Gamma_{2^N}\}$, can be regarded as $\wp(U) = \{S_1, S_2, \cdots, S_{2^N}\}$. The prediction error $\mathcal{E}(\mathcal{M}_i, \Gamma_i^{\text{tst}})$ can be regarded as $C(S_i)$. Therefore, our objective $\sum_{i=1}^{n} \mathcal{E}(\mathcal{M}_i, \Gamma_i^{\text{tst}})$ is equivalent to $\sum_{S_i \in \mathcal{K}}C(S_i)$ and thus is NP-hard.
\end{IEEEproof}

\begin{figure}[]
	\centering
	\includegraphics[width=\linewidth]{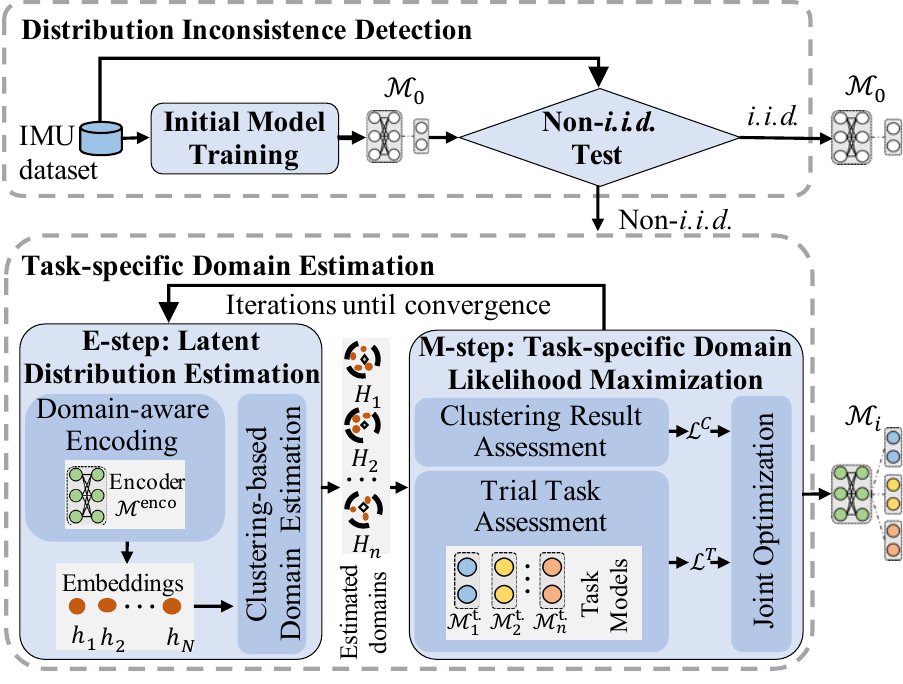}
	\caption{System architecture of Prism, where the IMU datasets are first detected whether they are non-\emph{i.i.d.} and the non-\emph{i.i.d.} IMU datasets is then partitioned for model training.}
	\label{fig:overview}
\end{figure}

\section{Design Overview}
The core idea of Prism is to effectively estimate latent domains regarding a specific perception task, embedded in a non-\emph{i.i.d.} IMU dataset, by partitioning data samples in an appropriate feature space. With estimated domains, versatile task models are trained together and downloaded to user devices for online prediction.
After downloading pre-trained models, FUP can be conducted locally on mobile devices.
To this end, as illustrated in Figure \ref{fig:overview}, Prism consists of three main parts as follows:

\textbf{Distribution Inconsistence Detection (DID).} Given a dataset $D$, DID trains an initial model $\mathcal{M}_0$ using all training samples, and then detects the inconsistency of data distributions in $D$, \emph{i.e.}, whether $D$ is a non-\emph{i.i.d.} dataset. For \emph{i.i.d.} datasets, Prism will directly utilize $\mathcal{M}_0$ for future online user perception. Otherwise, Prism will estimate task-specific domains for further model training.

\textbf{Task-specific Domain Estimation (TDE).} TDE is deployed on a cloud server, which estimates latent domains in the IMU dataset with an EM algorithm. Specifically, in the E-step, it first estimates the domains of data samples by clustering their features extracted with a backbone model $\mathcal{M}^{\text{enco}}$. Then, in the M-step, it maximizes the likelihood of estimated domains by considering the following two factors: 1) the feature similarity of each obtained domain; 2) the performance of a pack of $n$ task models $\mathcal{M}_i^{\text{task}}$ for $i \in [1, n]$, respectively trained and tested on such domains. These two steps repeat to optimize $\mathcal{M}^{\text{enco}}$ and all $\mathcal{M}_i^{\text{task}}$ until convergence.

\begin{figure}[]
	\centering
	\includegraphics[height=3cm]{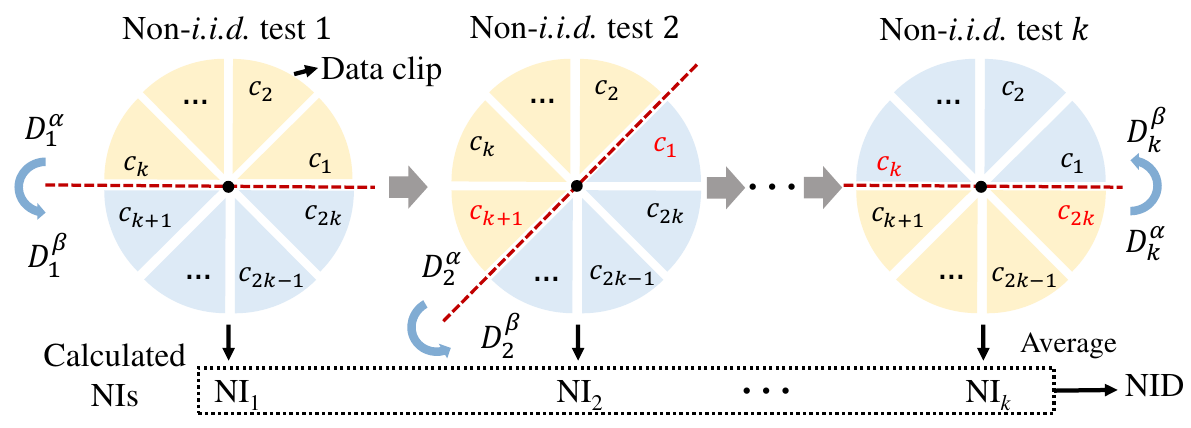}
	\caption{Illustration of NID calculation, where NIs are calculated and averaged as dataset $D$ is traversed by swapping clips between $D_i^{\alpha}$ and $D_i^{\beta}$.}
	\label{fig:ni_swap}
\end{figure}

\textbf{Online User Perception (OUP).}
Before conducting online user perception, all derived models in TDE, \emph{i.e.}, the $\mathcal{M}^{\text{enco}}$ and all $\mathcal{M}_i^{\text{task}}$ for $i \in [1, n]$, are downloaded to a mobile device. Each test data sample is first embedded into the same feature space using $\mathcal{M}^{\text{enco}}$. Then, the best-fit model classifier specific to the closest domain in the feature space will be selected for model inference. 

% \begin{figure}[t]
% 	\centering
% %	\vspace{-0.4cm}
% 	\includegraphics[width=0.9\linewidth]{figures/brief_show.pdf}
% 	\vspace{-0.2cm}
% 	\caption{Illustration of models involved in Prism, which consists of a unified feature extraction backbone $\mathcal{M}^{\text{enco}}$ and $n$ specific downstream task classifiers $\mathcal{M}_i^{\text{task}}$ for $i \in [1,n]$.}
% 	\vspace{-0.5cm}
% 	\label{fig:brief_show}
% \end{figure}

\section{Distribution Inconsistence Detection}
\subsection{Initial Model Training}

Given the available dataset $D$, we first train a model $\mathcal{M}_0$ for a specific perception task using all training samples. Specifically, $\mathcal{M}_0$ comprises a feature extraction backbone network $\mathcal{M}_0^{\text{enco}}$ and a task classifier $\mathcal{M}_0^{\text{task}}$. We train $\mathcal{M}_0$ with all training samples using the cross-entropy loss:
$\mathcal{L}^{\text{imt}} = - \frac{1}{N} \sum_{i=1}^{N} \sum_{j=1}^{C} y_{i,j} \log(\mathcal{M}_0^{\text{task}}(\mathcal{M}_0^{\text{enco}}(x_i))_j)$,
where $x_i$ denote the $i$-th data sample in dataset $D$; $C$ denotes the number of classes; $y_{i,j}$ denotes a binary label indicating whether $i$-th sample belongs to $j$-th label; and $N$ is the number of samples in dataset $D$.
%(\emph{i.e.}, $\mathcal{M}_0^{\text{task}}(\mathcal{M}_0^{\text{enco}}(x_i)) = \{\mathcal{M}_0^{\text{task}}(\mathcal{M}_0^{\text{enco}}(x_i))_j\}$ for $j \in [1,C]$)

\subsection{Non-\emph{i.i.d.} Test}

Given the dataset $D$ and the initial model $\mathcal{M}_0$, we examine the non-\emph{i.i.d.} level of $D$ regarding a particular perception task to determine whether latent domains should be identified. 
Specifically, we first partition $D$ randomly into $2k$ data clips, denoted as $c_i$, for $i \in [1, 2k]$, where $k$ denotes the pre-set epochs of non-\emph{i.i.d.} tests and $i$ denotes the index of rounds. Then, we group all clips into two sets, each with $k$ clips, denoted as $D_1^{\alpha}$ and $D_1^{\beta}$, respectively. For example, $D^{\alpha}_1 \gets \{c_1, c_2, \cdots, c_k\}$ and $D^{\beta}_1 \gets \{c_{k+1}, c_{k+2}, \cdots, c_{2k}\}$. 

The non-\emph{i.i.d.} index (NI) between two sub-datasets $D^{\alpha}_i$ and $D^{\beta}_i$ can be calculated as the norm of their features \cite{he2021towards}:
\begin{equation}
	\text{NI}_i = \frac{1}{C} \sum_{cls=1}^{C} \| \frac{\overline{\mathcal{M}^{\text{enco}}_0({[D^{\alpha \vphantom{\beta}}_i]}^{cls})}-\overline{\mathcal{M}^{\text{enco}}_0({[D^{\beta}_i]}^{cls})}}{\sigma (\mathcal{M}^{\text{enco}}_0([D]^{cls}))} \|_2,
\end{equation}
where ${[D^{\alpha}_i]}^{cls}$, ${[D^{\beta}_i]}^{cls}$ and $[D]^{cls}$ denotes the set of data samples in $D^{\alpha}_i$, $D^{\beta}_i$ and $D$ with class label $cls$, respectively; $\overline{(\cdot)}$ denotes the first order moment; $\sigma(\cdot)$ denotes the standard deviation used to normalize the scale of features; $\|\cdot\|_2$ denotes the L2-norm; $C$ denotes the number of classes to classify in a specific perception task. 

As illustrated in Figure \ref{fig:ni_swap}, we repeat the NI calculation between different pairs of $D^{\alpha}_i$ and $D^{\beta}_i$, constructed by exchanging $c_{i-1}$ in $D_{i-1}^{\alpha}$ and $c_{k+i-1}$ in $D_{i-1}^{\beta}$, for $i \in [2,k]$. 
For example, $D_2^\alpha \gets \{c_2, \cdots, c_k, c_{k+1}\}$ and $D^\beta_2 \gets \{c_{k+2}, \cdots, c_{2k}, c_1\}$, where $c_1$ in $D^\alpha_1$ and $c_{k+1}$ in $D^\beta_1$ are exchanged. After $k-1$ rounds of exchanges, $D^\alpha_1$ and $D^\beta_2$ are totally swapped, \emph{i.e.}, $D_k^\alpha = D_1^\beta$ and $D_k^\beta = D_1^\alpha$, which completes one traversal of dataset $D$. We define NID of dataset $D$ as the average of all $\text{NI}_i$ obtained in one traversal of $D$:
\begin{equation}
	\text{NID} = \frac{1}{k} \sum_{i=1}^k \text{NI}_i.
\end{equation}

\begin{figure}[]
	\centering
	\includegraphics[height=2.8cm]{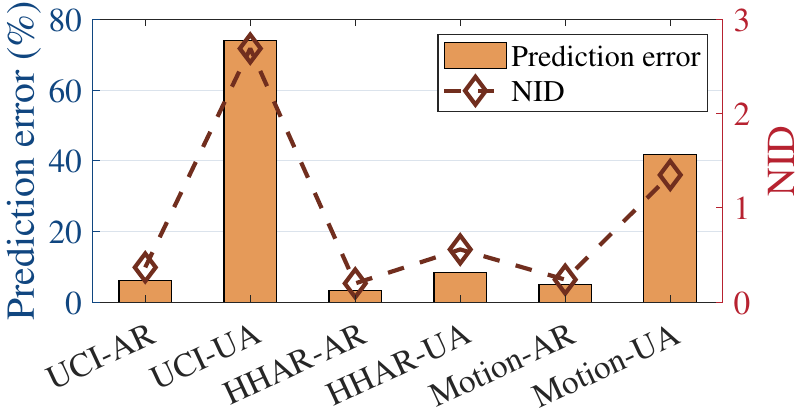}
	\caption{Non-\emph{i.i.d.} degree of a dataset (NID) vs perception prediction error, where a high prediction error is always with a high NID.}
	\label{fig:ni}
\end{figure}

Figure \ref{fig:ni} shows the prediction errors of a single CNN model and their corresponding NIDs of performing AR and UA tasks on three IMU user perception datasets, \emph{i.e.}, UCI, HHAR, and Motion. It can be seen that tasks with a high NID also have a high prediction error. Therefore, we consider dataset $D$ non-\emph{i.i.d.} for a task if its NID exceeds a certain threshold.

\section{Task-specific Domain Estimation}
%We identify latent task-specific domains and train corresponding task models at the same time using an EM algorithm. 

\subsection{Model Initialization}
\label{subsec:MI}
We first initialize all task models. Specifically, each $\mathcal{M}_{i}$ comprises a common backbone $\mathcal{M}^{\text{enco}}$ and $n$ domain-specific task classifiers $\mathcal{M}_i^{\text{task}}$ for $i \in [1,n]$, where $n$ represents the number of estimated domains. Each task classifier is trained for a particular domain to produce a prediction $\hat{y}$. The parameters of the pre-trained $\mathcal{M}_0$ are used to initialize all $\mathcal{M}_i$ for $i \in [1,n]$:
\begin{equation}
	\mathcal{M}^{\text{enco}} \gets \mathcal{M}_0^{\text{enco}},	
	\mathcal{M}_i^{\text{task}} \gets \mathcal{M}_0^{\text{task}} \text{ for } i \in [1,n].
\end{equation}

\subsection{E-step: Latent Distribution Estimation}
\label{subsec:LDM}

The E-step of Prism aims to estimate the task-specific domains in $D$. Specifically, we first utilize the encoder $\mathcal{M}^{\text{enco}}$ to encode $\{x_1, x_2, \cdots, x_N \}$ into $\{h_1, h_2, \cdots, h_N\}$ in a feature space. Then, we group $\{h_1, h_2, \cdots, h_N\}$ into $n$ domains $\{H_1, H_2, \cdots, H_n\}$ by clustering in the feature space through $k$-means. Each $H_i$ for $i \in [1,n]$ in the feature space corresponds to a latent domain $\Gamma_i$ in the data space, leading to a partition scheme $\hat{\mathcal{P}^*}$.

\subsection{M-step: Task-oriented Domain Likelihood Maximization}
\label{sec:dlm}

The parameters of backbone encoder $\mathcal{M}^{\text{enco}}$ are further optimized in a manner of gradient descent in M-step. In Prism, two assessments are designed to obtain the loss for optimization.

\subsubsection{Clustering Result Assessment} 

The clustering result assessment aims to assess whether the hidden feature space is well embedded so that the domains $\{\Gamma_1, \Gamma_2, \cdots, \Gamma_n\}$ are well divided in the feature space. The contrastive loss \cite{wang2021understanding} is used for clustering result assessment, which encourages similar pairs to be closer and dissimilar pairs to be farther apart in the feature space, which can be computed as follows:
\begin{equation}
\label{eq:c}
	\mathcal{L}^{C} = \frac{1}{2N} \sum_{n=1}^{N} [u \cdot d^2 + (1-u) \cdot \max(M - d, 0)^2],
\end{equation}
where $u$ denotes a binary label indicating whether two input samples belong to the same class ($u=1$) or not ($u=0$); $d$ denotes the distance between two samples in the feature space; $M$ denotes the contrastive margin, which is a hyper-parameter that determines the minimum distance for different-class samples. 

\subsubsection{Trial Task Assessment} 

To obtain the task-oriented loss, denoted as $\mathcal{L}^T$, the task classifiers $\mathcal{M}^{\text{task}}_i$ for $i \in [1,n]$ are jointly trained to access the quality of current partition scheme $\hat{\mathcal{P}^*}$.
Specifically, for each training sample $x_j$ and its corresponding domain index $m_j \in [1,n]$, we forward the features $h_j$ of $x_j$ with the $m_j$-th classifier. Then, the cross-entropy loss is used for the optimization of $\mathcal{M}^{\text{enco}}$ and $\mathcal{M}^{\text{task}}_i$ for $i \in [1,n]$:
\begin{equation}
\label{eq:t}
	\mathcal{L}^{T} = - \frac{1}{N} \sum_{j=1}^{N} \sum_{k=1}^{C} y_{j,k} \log(\mathcal{M}_{m_j}^{\text{task}}(\mathcal{M}^{\text{enco}}(x_j))_k).
\end{equation}

Finally, the total loss (denoted as $\mathcal{L}^{\text{tde}}$) for domain estimation is defined as the weighted sum of the contrastive loss $\mathcal{L}^C$ and the task-specific cross-entropy loss $\mathcal{L}^T$:
\begin{equation}
\label{eq:tde}
	\mathcal{L}^{\text{tde}} = \alpha \cdot \mathcal{L}^{C} + \mathcal{L}^{T},
\end{equation}
where $\alpha$ denotes a hyper-parameter of contrastive loss weight for the balance of $\mathcal{L}^{C}$ and $\mathcal{L}^{T}$.

\subsection{Theoretical Analyses}

\textbf{Convergence Analysis of Prism.} Prism can be proven to converge as follows.
\begin{IEEEproof}[Convergence of Prism]
    Denote the set of all parameters in $\mathcal{M}_i$ for $i \in [1,n]$ to be $\theta$. In Prism, we first estimate the domains $H_1, H_2, \cdots, H_n$ in E-step and then update the current parameters, denoted as $\theta^{(t)}$, to $\theta^{(t+1)}$ by minimizing the loss function $\mathcal{L}^{tde}$ shown in Equation \ref{eq:tde}. Therefore, to prove the convergence of Prism is to prove the convergence of $\mathcal{L}^{tde}$. To this end, we first prove the \emph{monotonicity} of $\mathcal{L}^{tde}$ during iteration and then prove the \emph{boundedness} of $\mathcal{L}^{tde}$.
    
\emph{Monotonicity.} The monotonicity of $\mathcal{L}^{tde}$ during iterations, \emph{i.e.}, $\mathcal{L}^{\text{tde}}(\theta^{(t+1)}) \leq \mathcal{L}^{\text{tde}}(\theta^{(t)})$ for each $t$, can be guaranteed in the M-step. Specifically, in M-step, we obtain $\theta^{(t+1)}$ by minimizing $\mathcal{L}^{tde}$, \emph{i.e.}, $\theta^{(t+1)} = \arg \min_\theta \mathcal{L}^{tde}(\theta^{(t)})$. As a result, we have $\mathcal{L}^{\text{tde}}(\theta^{(t+1)}) \leq \mathcal{L}^{\text{tde}}(\theta^{(t)})$.
    
    \emph{Boundedness.} We consider the custom loss function $\mathcal{L}^{\text{tde}} = \alpha \cdot \mathcal{L}^{C} + \mathcal{L}^{T}$, where $\mathcal{L}^{C}$ and $\mathcal{L}^{T}$ are shown in Equation \ref{eq:c} and Equation \ref{eq:t}, respectively. $\mathcal{L}^{\text{tde}}$ has a lower bound of 0 because both components of $\mathcal{L}^{\text{tde}}$, $\mathcal{L}^{C}$ and $\mathcal{L}^{T}$, have a lower bound of 0. For $\mathcal{L}^{C} = \frac{1}{2N} \sum_{n=1}^{N} [u \cdot d^2 + (1-u) \cdot \max(M - d, 0)^2]$, since both $d^2$ and $\max(M - d, 0)^2$ are non-negative, $\mathcal{L}^{C}$ is non-negative, and its minimum value is 0 when $d = 0$.
    For $\mathcal{L}^{T} = - \frac{1}{N} \sum_{j=1}^{N} \sum_{k=1}^{C} y_{j,k} \log(\mathcal{M}_{m_j}^{\text{task}}(\mathcal{M}^{\text{enco}}(x_j))_k)$, since $0 \leq \mathcal{M}_{m_j}^{\text{task}}(\mathcal{M}^{\text{enco}}(x_j))_k \leq 1$ and $\log(\mathcal{M}_{m_j}^{\text{task}}(\mathcal{M}^{\text{enco}}(x_j))_k) \leq 0$, $\mathcal{L}^{T}$ is non-negative, and its minimum value is 0 when $\mathcal{M}_{m_j}^{\text{task}}(\mathcal{M}^{\text{enco}}(x_j))_k = 1$ for the correct class $k$. Therefore, the combined loss function \(\mathcal{L}^{\text{tde}}\) is bounded below by 0, ensuring the boundedness of the loss function. 

    In conclusion, Prism is guaranteed to converge due to the monotonicity in each iteration and the bounded nature of the loss function $\mathcal{L}^{\text{tde}}$.
\end{IEEEproof}

\textbf{Computing Complexity Analysis of Training Prism.} The computational complexity of training Prism is linear.

\begin{IEEEproof}[Linear Complexity of Prism]
    Let $T_{\text{enco}}$ and $ T_{\text{task}}$ denote training time of one sample needed by $\mathcal{M}^{\text{enco}}$ and $\mathcal{M}^{\text{task}}$ for $i \in [1,n]$, respectively. Both $T_{\text{enco}}$ and $T_{\text{task}}$ are constant because of the nature of DNN \cite{song2017complexity}. Let $I_{\text{clus}}$ denote the iteration times of clustering.
    Since the overhead of the NID test is much smaller than model training, the computing time of Prism mainly consists of 3 parts: encoder training, domain clustering, and downstream task training. Thus, the overall complexity $\mathcal{O}(T_{\text{enco}} \cdot N + n \cdot I_{\text{clus}} \cdot N + T_{\text{task}} \cdot N) = \mathcal{O}((T_{\text{enco}} + n \cdot I_{\text{clus}} + T_{\text{task}}) \cdot N)$, where $n$ denotes the number of estimated domains. Note that all the coefficients of $N$ are constants. As a result, the overall time complexity is $\mathcal{O}(N)$.
\end{IEEEproof}

Although the coefficient of linear complexity is large, compared to the original NP-hard problem, the complexity is greatly reduced and is acceptable for training on the cloud.

\section{Evaluation}
\label{sec:eval}

\subsection{Methodology}

\subsubsection{Datasets} 
\label{sec:dataset}

We consider the following user perception datasets:

\begin{itemize}[leftmargin=*]
	\item \textbf{UCI} \cite{reyes2016transition}: UCI is a publicly available dataset containing accelerometer and gyroscope readings from a Samsung Galaxy S II smartphone carried by 30 subjects when performing six activities, \emph{i.e.}, standing, sitting, lying, walking, going downstairs, and going upstairs. The data sampling rate is 50 Hz. We slice each sensor trace into non-overlapping segments of 300 samples and filter out segments with multiple activity labels, leading to a set $D$ of 2,088 IMU segments.
	\item \textbf{HHAR} \cite{stisen2015smart}: HHAR is a publicly available dataset consisting of accelerometer and gyroscope readings collected from 6 types of mobile phones (3 models of Samsung Galaxy and 1 model of LG). The smartphones are worn around the waist by 9 users performing 6 different activities (\emph{biking, sitting, standing, walking, upstairs, and downstairs}). The sampling rates of HHAR are 100 - 200 Hz.
	\item \textbf{Motion} \cite{malekzadeh2019mobile}: Motion is a publicly available dataset of accelerometer and gyroscope readings collected from a smartphone (iPhone 6s) worn by 24 subjects during various daily activities. The data is collected with the smartphone in the front pockets of the subjects. Motion covers 6 different activities (\emph{downstairs, upstairs, walking, jogging, sitting, and standing}) at a sampling rate of 50 Hz.
\end{itemize} 

We down-sample the IMU data to 20 Hz and slice the data with a window length of 120, each with a window of 6s. We omit those samples with multiple inconsistent labels and obtain a dataset of 2088, 5434, and 9166 original IMU samples for UCI, HHAR, and Motion, respectively. We normalize the recordings as follows: $a_i = \frac{a_i}{g},~i\in\{x, y, z\}$, where $a_i$ denotes the accelerometer readings in the $i$-axis, respectively; $g$ denotes the universal gravitational constant. Data samples are then shuffled and divided into training sets, validation sets, and testing sets with a ratio of 6:2:2. 

\subsubsection{Implementation} 
\label{sec:impl}

We implement the offline domain estimation and model training part on a cloud server equipped with 256GB DRAM and 4 Nvidia 3090 GPUs. 
% The model backbone $\mathcal{M}^{\text{enco}}$ is trained based on ConvLSTM for its better ability to handle time-series data like IMU data \cite{Qian2022, li2021cross}. 
Every downstream task classifier $\mathcal{M}_i^{\text{task}}$ for $i \in [1,n]$ is comprised of an MLP \cite{taud2018multilayer}. Empirically, Prism can converge within 200 epochs. As a result, Prism conducts EM iterations for 200 epochs and the model with the best performance on the validation set is selected for further testing.

We implement the online inference part on 6 typical mobile devices, \emph{i.e.}, Honor X40, Vivo X27, Mi 6, Pixel 3 XL, Huawei Mate 40 Pro, and iPhone 14 Plus. The hardware configurations are shown in Table~\ref{tab:hardware}. ONNX \cite{bai2019} is used to convert models for cross-platform deployment. Well-trained model $\mathcal{M}_{i}$ is offline downloaded from the cloud server to mobile devices. TVM \cite{chen2018tvm} is used for the model acceleration on mobile devices.

\begin{table}[b]
	\caption{Prism is implemented on 6 different types of mobile phones with distinct hardware configurations.}
	\label{tab:hardware}
	\centering
	\scalebox{0.9}{
		\begin{tabular}{c c c c c}
			\toprule
			Phone              & SoC            & GPU           & Memory & Disk  \\ \hline
			Honor X40          & Dimensity 1300 & Mali-G77 MC9  & 12GB   & 256GB \\ 
			Vivo X27           & Snapdragon 710 & Adreno 616    & 8GB    & 256GB \\ 
			Mi 6               & Snapdragon 835 & Adreno 540    & 6GB    & 64GB  \\
			Pixel 3 XL         & Snapdragon 845 & Adreno 630    & 4GB    & 128GB \\
			Mate 40 Pro & Kirin 9000     & Mali-G78 MP24 & 8GB    & 256GB \\ 
			iPhone 14 Plus     & A15            & 5-core GPU    & 6GB    & 256GB \\
			\bottomrule
		\end{tabular}
	}
\end{table}

\subsubsection{Candidate Methods} 
We compare Prism with the following candidate methods:
\begin{itemize}
	\item \textbf{Training one single model ($\mathcal{P}^0$)}: A single model is trained with all available samples and is tested for all test samples.
	\item \textbf{Semantic partition ($\mathcal{P}^{\text{sem}}$)} \cite{zheng2022towards}: Models are trained on domains defined by semantic attributes. During online prediction, the task model with the same semantic attribute is selected for use.
	\item \textbf{Clustering in the data space ($\mathcal{P}^{\text{CD}}$)} \cite{zheng2022life}: Models are trained on domains defined by clustering training samples in the original data space. During online prediction, the downstream task model whose mean of all original data is closest to the data of the test sample is selected for use.
	\item \textbf{Clustering in the feature space ($\mathcal{P}^{\text{CF}}$)} \cite{lu2023out}: Models are trained on domains defined by deep clustering \cite{caron2018deep} on training samples. During online prediction, the downstream task model whose mean of all features is closest to that of the test sample is selected.
\end{itemize}

For all the candidate methods, we consider two model training schemes as follows:
\begin{itemize}
	\item {\textbf{Training one single model using domain generalization (DG):}} One single model, denoted as $\mathcal{M}_0^{\text{DG}}$, is trained by aligning data samples of the same label in each $\Gamma_i$ in the feature space \cite{lu2023out}.
	\item {\textbf{Training individual models using domain adaptation (DA):}} An Individual model, denoted as $\mathcal{M}_i^{\text{DA}}$, is trained by fine-tuning $\mathcal{M}_0$ on each $\Gamma_i$ \cite{zheng2022towards}.
\end{itemize}

\subsubsection{Tasks and Metrics}

We compare all candidate methods on typical user perception tasks of Activity Recognition (AR) and User Authentication (UA). In the AR task, models are trained to recognize human activities (\emph{e.g.}, standing, lying, or walking) with IMU data. In the UA task, models are trained to recognize human IDs (\emph{e.g.}, User 1 and User 2). As the considered tasks are classification tasks, we adopt accuracy (Acc) and F1 score (F1) for performance comparison. Acc is defined as the proportion of correctly predicted samples to the total number of test samples and F1 is defined as $\text{F1} = \frac{1}{N_C} \sum_{i=1}^{N_C} \frac{2 \cdot p_i \cdot r_i}{p_i+r_i}$, where $p_i$ and $r_i$ denote the precision and recall of the $i$-th class, respectively, and $N_C$ denotes the number of all classes.

\subsection{Hyper-parameters Selection}

\begin{figure}[]
	\centering
	\subfigure[Weight of the contrastive loss $\mathcal{L}^C$]{
		\begin{minipage}[t]{0.140\textwidth}
			\includegraphics[height=2.2cm]{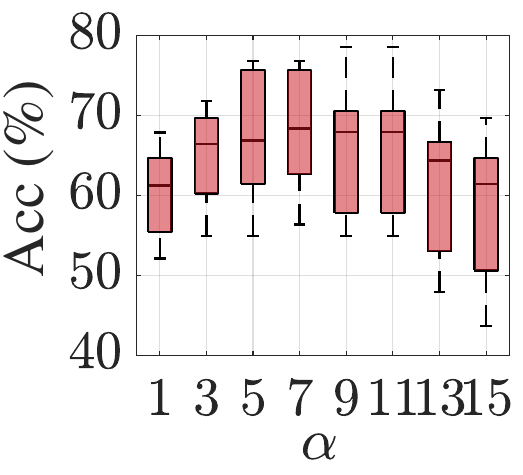}
		\end{minipage}
		\label{fig:cl_weight}
	}
	%\hspace{0.1cm}
	\subfigure[Number of estimated domains]{
		\begin{minipage}[t]{0.140\textwidth}
			\includegraphics[height=2.2cm]{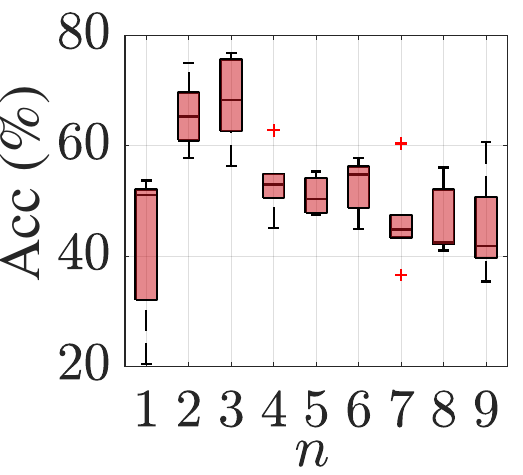}
		\end{minipage}
		\label{fig:cl_num}
	}
	%\hspace{0.11cm}
	\subfigure[Contrastive margin]{
		\begin{minipage}[t]{0.140\textwidth}
			\includegraphics[height=2.2cm]{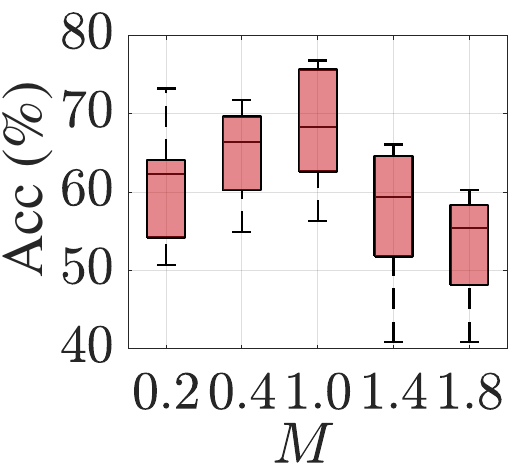}
		\end{minipage}
		\label{fig:margin}
	}
	\caption{Hyper-parameters selection results across 6 testing sets $\Gamma_i^{\text{tst}}$ for $i \in [1,6]$ on the UA task of UCI dataset.}
	\label{fig:hyper}
\end{figure}

\begin{table*}[]
	\caption{Overall performance of Prism and all other candidate methods, where Prism outwits other candidate methods on various tasks.}
	\label{tab:overall}
	\centering
	\scalebox{1}{
		\begin{tabular}{cc|cccc|cccc|cccc}
\hline
\multicolumn{2}{c|}{Dataset}                                          & \multicolumn{4}{c|}{UCI}                                                                                                                                                                                                                                                                                                                                             & \multicolumn{4}{c|}{HHAR}                                                                                                                                                                                                                                                                                                                                            & \multicolumn{4}{c}{Motion}                                                                                                                                                                                                                                                                                                                                           \\ \hline
\multicolumn{2}{c|}{Task}                                             & \multicolumn{2}{c|}{AR}                                                                                                                                                                     & \multicolumn{2}{c|}{UA}                                                                                                                                                & \multicolumn{2}{c|}{AR}                                                                                                                                                                     & \multicolumn{2}{c|}{UA}                                                                                                                                                & \multicolumn{2}{c|}{AR}                                                                                                                                                                     & \multicolumn{2}{c}{UA}                                                                                                                                                 \\ \hline
\multicolumn{2}{c|}{NID}                                              & \multicolumn{2}{c|}{0.37}                                                                                                                                                                   & \multicolumn{2}{c|}{\textbf{2.69}}                                                                                                                                     & \multicolumn{2}{c|}{0.20}                                                                                                                                                                   & \multicolumn{2}{c|}{0.56}                                                                                                                                              & \multicolumn{2}{c|}{0.24}                                                                                                                                                                   & \multicolumn{2}{c}{\textbf{1.35}}                                                                                                                                      \\ \hline
\multicolumn{2}{c|}{Metric (\%)}                                      & \multicolumn{1}{c|}{Acc}                                                                     & \multicolumn{1}{c|}{F1}                                                                      & \multicolumn{1}{c|}{Acc}                                                                     & F1                                                                      & \multicolumn{1}{c|}{Acc}                                                                     & \multicolumn{1}{c|}{F1}                                                                      & \multicolumn{1}{c|}{Acc}                                                                     & F1                                                                      & \multicolumn{1}{c|}{Acc}                                                                     & \multicolumn{1}{c|}{F1}                                                                      & \multicolumn{1}{c|}{Acc}                                                                     & F1                                                                      \\ \hline
\multicolumn{2}{c|}{$\mathcal{P}^0$}                                  & \multicolumn{1}{c|}{\color{blue} 93.06}                                     & \multicolumn{1}{c|}{\color{blue} 93.53}                                     & \multicolumn{1}{c|}{42.26}                                                                   & 40.11                                                                   & \multicolumn{1}{c|}{\color{blue} 98.26}                                     & \multicolumn{1}{c|}{\color{blue} 98.15}                                     & \multicolumn{1}{c|}{\color{blue} 95.69}                                     & \color{blue} 95.71                                     & \multicolumn{1}{c|}{\color{blue} 98.53}                                     & \multicolumn{1}{c|}{\color{blue} 97.81}                                     & \multicolumn{1}{c|}{75.19}                                                                   & 74.46                                                                   \\ \hline
\multicolumn{1}{c|}{\multirow{3}{*}{DG}} & $\mathcal{P}^{\text{sem}}$ & \multicolumn{1}{c|}{72.73}                                                                   & \multicolumn{1}{c|}{67.78}                                                                   & \multicolumn{1}{c|}{43.54}                                                                   & 38.11                                                                   & \multicolumn{1}{c|}{76.05}                                                                   & \multicolumn{1}{c|}{75.78}                                                                   & \multicolumn{1}{c|}{64.99}                                                                   & 58.47                                                                   & \multicolumn{1}{c|}{86.48}                                                                   & \multicolumn{1}{c|}{78.48}                                                                   & \multicolumn{1}{c|}{64.46}                                                                   & 61.45                                                                   \\
\multicolumn{1}{c|}{}                    & $\mathcal{P}^{\text{CD}}$  & \multicolumn{1}{c|}{82.06}                                                                   & \multicolumn{1}{c|}{78.85}                                                                   & \multicolumn{1}{c|}{36.68}                                                                   & 30.94                                                                   & \multicolumn{1}{c|}{73.59}                                                                   & \multicolumn{1}{c|}{73.24}                                                                   & \multicolumn{1}{c|}{72.32}                                                                   & 70.61                                                                   & \multicolumn{1}{c|}{91.69}                                                                   & \multicolumn{1}{c|}{85.72}                                                                   & \multicolumn{1}{c|}{67.51}                                                                   & 63.80                                                                   \\
\multicolumn{1}{c|}{}                    & $\mathcal{P}^{\text{CF}}$  & \multicolumn{1}{c|}{72.73}                                                                   & \multicolumn{1}{c|}{67.78}                                                                   & \multicolumn{1}{c|}{43.54}                                                                   & 38.11                                                                   & \multicolumn{1}{c|}{76.05}                                                                   & \multicolumn{1}{c|}{75.78}                                                                   & \multicolumn{1}{c|}{64.99}                                                                   & 58.47                                                                   & \multicolumn{1}{c|}{86.48}                                                                   & \multicolumn{1}{c|}{78.48}                                                                   & \multicolumn{1}{c|}{64.46}                                                                   & 61.45                                                                   \\ \hline
\multicolumn{1}{c|}{\multirow{3}{*}{DA}} & $\mathcal{P}^{\text{sem}}$ & \multicolumn{1}{c|}{81.9}                                                                    & \multicolumn{1}{c|}{77.19}                                                                   & \multicolumn{1}{c|}{40.75}                                                                   & 33.95                                                                   & \multicolumn{1}{c|}{57.16}                                                                   & \multicolumn{1}{c|}{55.26}                                                                   & \multicolumn{1}{c|}{54.33}                                                                   & 46.85                                                                   & \multicolumn{1}{c|}{86.81}                                                                   & \multicolumn{1}{c|}{82.93}                                                                   & \multicolumn{1}{c|}{\color{blue} 79.82}                                     & \color{blue} 77.67                                     \\
\multicolumn{1}{c|}{}                    & $\mathcal{P}^{\text{CD}}$  & \multicolumn{1}{c|}{87.88}                                                                   & \multicolumn{1}{c|}{85.10}                                                                   & \multicolumn{1}{c|}{\color{blue} 49.28}                                     & \color{blue} 44.16                                     & \multicolumn{1}{c|}{62.59}                                                                   & \multicolumn{1}{c|}{61.01}                                                                   & \multicolumn{1}{c|}{70.47}                                                                   & 68.14                                                                   & \multicolumn{1}{c|}{70.77}                                                                   & \multicolumn{1}{c|}{60.68}                                                                   & \multicolumn{1}{c|}{70.01}                                                                   & 67.63                                                                   \\
\multicolumn{1}{c|}{}                    & $\mathcal{P}^{\text{CF}}$  & \multicolumn{1}{c|}{86.05}                                                                   & \multicolumn{1}{c|}{81.93}                                                                   & \multicolumn{1}{c|}{40.19}                                                                   & 34.67                                                                   & \multicolumn{1}{c|}{90.77}                                                                   & \multicolumn{1}{c|}{90.12}                                                                   & \multicolumn{1}{c|}{62.83}                                                                   & 58.66                                                                   & \multicolumn{1}{c|}{79.46}                                                                   & \multicolumn{1}{c|}{76.28}                                                                   & \multicolumn{1}{c|}{63.14}                                                                   & 60.21                                                                   \\ \hline
\multicolumn{2}{c|}{$\mathcal{P}^{\text{pri}}$}                       & \multicolumn{1}{c|}{\color{green!60!black} \textbf{95.93}} & \multicolumn{1}{c|}{\color{green!60!black} \textbf{96.18}} & \multicolumn{1}{c|}{\color{green!60!black} \textbf{58.61}} & \color{green!60!black} \textbf{56.90} & \multicolumn{1}{c|}{\color{green!60!black} \textbf{99.16}} & \multicolumn{1}{c|}{\color{green!60!black} \textbf{99.07}} & \multicolumn{1}{c|}{\color{green!60!black} \textbf{96.73}} & \color{green!60!black} \textbf{96.69} & \multicolumn{1}{c|}{\color{green!60!black} \textbf{98.75}} & \multicolumn{1}{c|}{\color{green!60!black} \textbf{98.16}} & \multicolumn{1}{c|}{\color{green!60!black} \textbf{82.98}} & \color{green!60!black} \textbf{82.67} \\ \hline
\multicolumn{2}{c|}{Gain over $\mathcal{P}^0$}                        & \multicolumn{1}{c|}{+2.87}                                                                   & \multicolumn{1}{c|}{+2.65}                                                                   & \multicolumn{1}{c|}{+16.35}                                                                  & +16.79                                                                  & \multicolumn{1}{c|}{+0.90}                                                                   & \multicolumn{1}{c|}{+0.92}                                                                   & \multicolumn{1}{c|}{+1.04}                                                                   & +0.98                                                                   & \multicolumn{1}{c|}{+0.22}                                                                   & \multicolumn{1}{c|}{+0.35}                                                                   & \multicolumn{1}{c|}{+7.79}                                                                   & +8.21                                                                   \\ \hline
\end{tabular}
	}
\end{table*}

We first investigate the selection of hyper-parameters, \emph{i.e.}, weight of contrastive loss $\alpha$, number of estimated domains $n$, and contrastive margin $M$. We conduct hyper-parameters selection on UA task of UCI dataset, which is a representative non-\emph{i.i.d.} task, with an NID of 2.69. In order to present the detailed testing results, we partition the testing set of UCI based on prior semantic attributes, \emph{i.e.}, activities, and obtain 6 testing sets, denoted as $\Gamma_i^{\text{tst}}$ for $i \in [1,6]$ for evaluation. 

\textbf{Weight of Contrastive Loss $\alpha$.}
Figure \ref{fig:cl_weight} shows the box plot of FUP accuracy on $\Gamma_i^{\text{tst}}$ for $i \in [1,6]$ with different contrastive loss $\alpha$ in Prism. It can be seen that both a low $\alpha$ and a high $\alpha$ result in a relatively low inference accuracy. This is because a low $\alpha$ makes Prism degenerate to the methods without partition while a high $\alpha$ will influence the optimization with $\mathcal{L}^{T}$, which makes the training of downstream tasks under-fitting. As a result, in the following experiments, we choose a moderate value for $\alpha$ for its best performance.

\textbf{Number of Estimated Domains $n$.}
Figure~\ref{fig:cl_num} shows the box plot of FUP accuracy of Prism on $\Gamma_i^{\text{tst}}$ for $i \in [1,6]$ with different domain estimating numbers $n$. It can be seen that the test accuracy increases quickly at first with $n$ increasing. This shows the efficacy of domain estimation on the non-\emph{i.i.d.} tasks. We can also see that the accuracy drops with the increase of $n$. This is because \emph{i.i.d.} domains may also be partitioned with a large $n$, resulting in a sub-optimal FUP performance. 
As a result, in the following experiments, we choose an intermediate value of $n$ for its competitive performance.

\textbf{Contrastive Margin $M$.}
Figure \ref{fig:margin} shows the box plot of FUP accuracy of Prism on $\Gamma_i^{\text{tst}}$ for $i \in [1,6]$ with different contrastive margin $M$. It can be seen that the accuracy first increase and then drop with the increase of $M$. This is because a too small $M$ will make $\mathcal{L}^C$ ignore some key samples while a too high $M$ will make $\mathcal{L}^C$ focus on the hard samples. Both of the above two cases will result in a worse FUP performance. As a result, we choose a moderate value for $M$ (\emph{e.g.}, $M=1.0$ for UA task on UCI dataset) for better performance.

\subsection{Overall FUP Performance}

\label{sec:overall}

In this experiment, we investigate the performance of all candidate methods on FUP user perception tasks. The task-specific domain estimation (TDE) module is conducted for all datasets to evaluate its effect on datasets with various NIDs. The models are trained with DCNN, GRU, and Transformer, respectively. Table \ref{tab:overall} shows the average accuracy and F1 score of candidate methods of the models for all tasks.

\subsubsection{Performance Comparison}

It can be seen that Prism outwits other methods over all tasks on all datasets. Prism outperforms the traditional method $\mathcal{P}^0$ on tasks with high non-\emph{i.i.d.} degree (NID) by up to 38.69\% and 41.86\% relatively in terms of accuracy and F1 score, respectively. This demonstrates that FUP accuracy will be better for non-\emph{i.i.d.} datasets if we consider its non-\emph{i.i.d.} issue. On tasks with low NID, where the performance of $\mathcal{P}^0$ is good enough, Prism can also slightly outperform $\mathcal{P}^0$ and other candidate methods. This shows the adaptability of Prism on both tasks with high and low NID.

\subsubsection{Performance of Semantic Partition}

It is a common practice to partition a dataset based on semantic attributes (\emph{i.e.}, $\mathcal{P}^{\text{sem}}$). It can be seen that Prism can outperform $\mathcal{P}^{\text{sem}}$ with both DA and DG training schemes. The average prediction accuracy of $\mathcal{P}^{\text{sem}}$ is even lower than $\mathcal{P}^0$ by over 20\%. This is because the semantic attributes are task-agnostic and may not always work well on all tasks. $\mathcal{P}^{\text{sem}}$ may work well on some tasks. For example, on UA task of Motion dataset, $\mathcal{P}^{\text{sem}}$ with DG training scheme outperforms $\mathcal{P}^0$ on accuracy by 3.03\%. However, $\mathcal{P}^{\text{sem}}$ can not work well in most cases, which indicates its limitation.

\subsubsection{Performance of Domain Partition based on Clustering}

A na\"ive way for domain partition is clustering data samples in the dataset unsupervisedly on data space (\emph{i.e.}, $\mathcal{P}^{\text{CD}}$) or feature space(\emph{i.e.}, $\mathcal{P}^{\text{CF}}$). We can see that Prism outperforms both $\mathcal{P}^{\text{CD}}$ and $\mathcal{P}^{\text{CF}}$ on all tasks. This is because unsupervised clustering on data samples is also task-agnostic and remains unstable in performance.
Moreover, it can be seen that Prism outperforms the state-of-the-art DG methods \cite{lu2023out} (\emph{i.e.}, $\mathcal{P}^{\text{CF}}$ with DG training scheme) in terms of accuracy and F1 score by 41.94 \% and 46.53 \%, respectively. This is because FUP problem is different from domain generalization (DG) problem. FUP focuses on flexible model adaptation between seen domains while DG focuses on model generalization to unseen domains. As a result, DG can not work well on FUP problem. We will apply DA for model training in the following experiments.

\begin{table*}[]
	\caption{Average FUP performance on large-scale non-\emph{i.i.d.} SHL dataset, where Prism outperforms all the candidate methods on average. Each model is trained with domain adaptation for the superior performance of this training scheme on FUP problem.}
	\label{tab:overall_shl}
        \centering
	\scalebox{1}{
		\begin{threeparttable}
			\begin{tabular}{c|cccc|cccc|cccc|cccc}
\hline
Model                      & \multicolumn{4}{c|}{DCNN}                                                          & \multicolumn{4}{c|}{GRU}                                                           & \multicolumn{4}{c|}{LIMU-CNN}                                                      & \multicolumn{4}{c}{LIMU-GRU}                                                      \\ \hline
Task                       & \multicolumn{2}{c|}{AR}                            & \multicolumn{2}{c|}{UA}       & \multicolumn{2}{c|}{AR}                            & \multicolumn{2}{c|}{UA}       & \multicolumn{2}{c|}{AR}                            & \multicolumn{2}{c|}{UA}       & \multicolumn{2}{c|}{AR}                            & \multicolumn{2}{c}{UA}       \\ \hline
Metric (\%)                & Acc           & \multicolumn{1}{c|}{F1}            & Acc           & F1            & Acc           & \multicolumn{1}{c|}{F1}            & Acc           & F1            & Acc           & \multicolumn{1}{c|}{F1}            & Acc           & F1            & Acc           & \multicolumn{1}{c|}{F1}            & Acc           & F1            \\ \hline
$\mathcal{P}^0$            & 75.2          & \multicolumn{1}{c|}{76.1}          & 85.4          & 85.1          & \color{blue} 75.7          & \multicolumn{1}{c|}{\color{blue} 77.5}          & \color{blue} 91.0          & \color{blue} 91.0          & 77.0          & \multicolumn{1}{c|}{77.2}          & 82.5          & 82.3          & 78.4          & \multicolumn{1}{c|}{\color{blue} 80.0}          & 86.9          & 86.7          \\
$\mathcal{P}^{\text{CF}}$  & \color{blue} 81.2          & \multicolumn{1}{c|}{\color{blue} 82.1}          & 89.7          & 89.7          & 73.2          & \multicolumn{1}{c|}{75.0}          & 87.6          & 87.5          & 78.7          & \multicolumn{1}{c|}{79.3}          & 87.4          & 87.2          & 78.8          & \multicolumn{1}{c|}{79.8}          & \color{blue} 90.5          & \color{blue} 90.4          \\
$\mathcal{P}^{\text{CD}}$  & 81.0          & \multicolumn{1}{c|}{81.7}          & 90.2          & 90.1          & 72.9          & \multicolumn{1}{c|}{73.7}          & 88.0          & 87.8          & \color{blue} 79.0          & \multicolumn{1}{c|}{\color{blue} 79.5}          & 86.4          & 86.3          & \color{blue} 78.9          & \multicolumn{1}{c|}{\color{blue} 80.0}          & 90.4          & 90.3          \\
$\mathcal{P}^{\text{SR}}$  & 71.2          & \multicolumn{1}{c|}{72.3}          & 76.2          & 76.0          & 64.2          & \multicolumn{1}{c|}{65.2}          & 79.3          & 79.0          & 65.3          & \multicolumn{1}{c|}{64.1}          & 77.5          & 77.3          & 66.8          & \multicolumn{1}{c|}{65.2}          & 65.2          & 78.4          \\
$\mathcal{P}^{\text{sem}}$ & 78.8          & \multicolumn{1}{c|}{79.9}          & \color{blue} 91.8          & \color{blue} 91.7          & 71.1          & \multicolumn{1}{c|}{72.1}          & 89.1          & 89.0          & 74.4          & \multicolumn{1}{c|}{73.7}          & \color{green!60!black} \textbf{88.1} & \color{green!60!black} \textbf{88.0} & 75.0          & \multicolumn{1}{c|}{74.9}          & 89.5          & 89.0          \\
$\mathcal{P}^{\text{pri}}$ & \color{green!60!black} \textbf{84.7} & \multicolumn{1}{c|}{\color{green!60!black} \textbf{85.4}} & \color{green!60!black} \textbf{92.5} & \color{green!60!black} \textbf{92.4} & \color{green!60!black} \textbf{78.8} & \multicolumn{1}{c|}{\color{green!60!black} \textbf{80.2}} & \color{green!60!black} \textbf{91.5} & \color{green!60!black} \textbf{91.5} & \color{green!60!black} \textbf{79.2} & \multicolumn{1}{c|}{\color{green!60!black} \textbf{80.2}} & \color{blue} 88.0          & \color{blue} 87.8          & \color{green!60!black} \textbf{82.9} & \multicolumn{1}{c|}{\color{green!60!black} \textbf{83.9}} & \color{green!60!black} \textbf{91.0} & \color{green!60!black} \textbf{90.9} \\ \hline
Gain over $\mathcal{P}^0$  & +9.5          & \multicolumn{1}{c|}{+9.3}          & +7.1          & +7.3          & +3.1          & \multicolumn{1}{c|}{+2.7}          & +0.5          & +0.5          & +2.2          & \multicolumn{1}{c|}{+3.0}          & +5.5          & +5.5          & +4.5          & \multicolumn{1}{c|}{+3.9}          & +4.1          & +4.2          \\ \hline
\end{tabular}
\begin{tablenotes}
    \footnotesize
    \item[1] In fact, accurate semantic information is lacking in real-world testing. As a result, $\mathcal{P}^{\text{sem}}$ is unrealistic and we add a realistic alternative of $\mathcal{P}^{\text{sem}}$ named $\mathcal{P}^{\text{SR}}$ (semantic partition in the real-world setting), which utilizes a neural network for semantic information prediction in real time.
    % \item[2] The scale of NID also differs between models because of the different data scale of different encoder $\mathcal{M}^{\text{enco}}$. In general, the data scle of CNN-based encoder (\emph{e.g.}, DCNN and LIMU-CNN) is lager than that of RNN-based encoder (\emph{e.g.}, GRU and LIMU-GRU				).
\end{tablenotes}
		\end{threeparttable}
	}
\end{table*}

\begin{figure*}[]
	\centering
	\subfigure[AR: Accuracy]{
		\begin{minipage}[b]{0.22\textwidth}
			\includegraphics[width=1\linewidth]{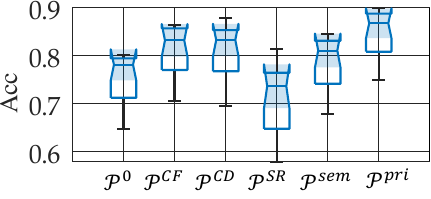}
		\end{minipage}
		\label{fig:dcnn_acc}
	}
	\subfigure[AR: F1 score]{
		\begin{minipage}[b]{0.22\textwidth}
			\includegraphics[width=1\linewidth]{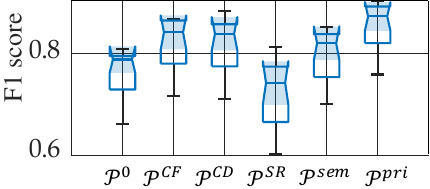}
		\end{minipage}
		\label{fig:dcnn_f1}
	}
	\subfigure[UA: Accuracy]{
		\begin{minipage}[b]{0.22\textwidth}
			\includegraphics[width=1\linewidth]{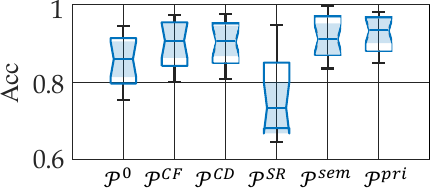}
		\end{minipage}
		\label{fig:dcnn_acc_ua}
	}
	\subfigure[UA: F1 score]{
		\begin{minipage}[b]{0.22\textwidth}
			\includegraphics[width=1\linewidth]{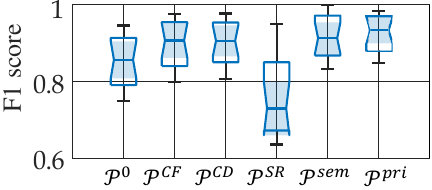}
		\end{minipage}
		\label{fig:dcnn_f1_ua}
	}
	\caption{Boxplots of FUP performance based on DCNN, where Prism can outperform all other candidate methods and achieve the best performance among all the base models. We denote the partition of Prism as $\mathcal{P}^{\text{pri}}$.}
	\label{fig:dcnn_fine}
\end{figure*}

\begin{figure*}[]
	\centering
	\subfigure[AR: Accuracy]{
		\begin{minipage}[b]{0.22\textwidth}
			\includegraphics[width=1\linewidth]{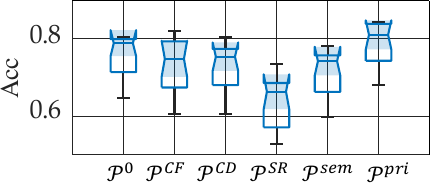}
		\end{minipage}
		\label{fig:gru_acc}
	}
	\subfigure[AR: F1 score]{
		\begin{minipage}[b]{0.22\textwidth}
			\includegraphics[width=1\linewidth]{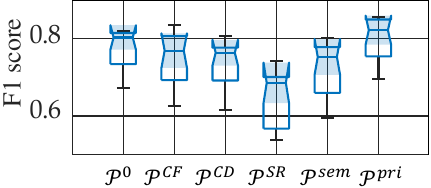}
		\end{minipage}
		\label{fig:gru_f1}
	}
	\subfigure[UA: Accuracy]{
		\begin{minipage}[b]{0.22\textwidth}
			\includegraphics[width=1\linewidth]{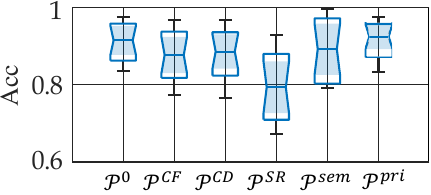}
		\end{minipage}
		\label{fig:gru_acc_ua}
	}
	\subfigure[UA: F1 score]{
		\begin{minipage}[b]{0.22\textwidth}
			\includegraphics[width=1\linewidth]{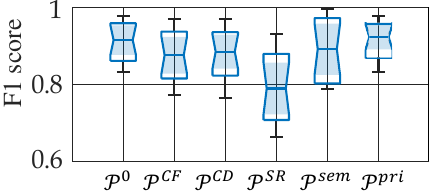}
		\end{minipage}
		\label{fig:gru_f1_ua}
	}
	\caption{Boxplots of FUP performance based on GRU, where Prism can outperform all other candidate methods. Note that on this base model, $\mathcal{P}^0$ works second best.}
	\label{fig:gru_fine}
\end{figure*}

\begin{figure*}[]
	\centering
	\subfigure[AR: Accuracy]{
		\begin{minipage}[b]{0.22\textwidth}
			\includegraphics[width=1\linewidth]{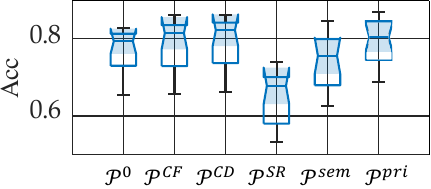}
		\end{minipage}
		\label{fig:limu_cnn_acc}
	}
	\subfigure[AR: F1 score]{
		\begin{minipage}[b]{0.22\textwidth}
			\includegraphics[width=1\linewidth]{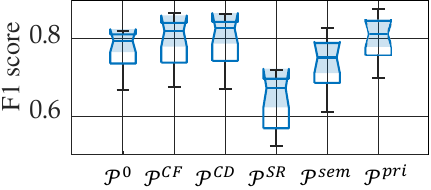}
		\end{minipage}
		\label{fig:limu_cnn_f1}
	}
	\subfigure[UA: Accuracy]{
		\begin{minipage}[b]{0.22\textwidth}
			\includegraphics[width=1\linewidth]{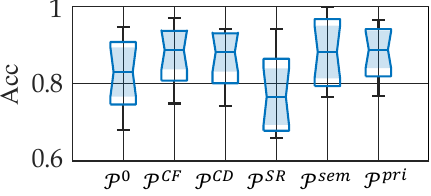}
		\end{minipage}
		\label{fig:limu_cnn_acc_ua}
	}
	\subfigure[UA: F1 score]{
		\begin{minipage}[b]{0.22\textwidth}
			\includegraphics[width=1\linewidth]{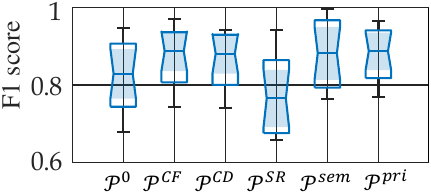}
		\end{minipage}
		\label{fig:limu_cnn_f1_ua}
	}
	\caption{Boxplots of FUP performance based on LIMU-CNN. Surprisingly, other partition-based methods also work well on this base model owing to the strong generalization ability of the pre-trained model LIMU. However, Prism is also comparable with the best methods.}
	\label{fig:limu_cnn_fine}
\end{figure*}

\begin{figure*}[]
	\centering
	\subfigure[AR: Accuracy]{
		\begin{minipage}[b]{0.22\textwidth}
			\includegraphics[width=1\linewidth]{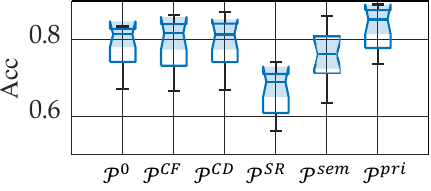}
		\end{minipage}
		\label{fig:limu_gru_acc}
	}
	\subfigure[AR: F1 score]{
		\begin{minipage}[b]{0.22\textwidth}
			\includegraphics[width=1\linewidth]{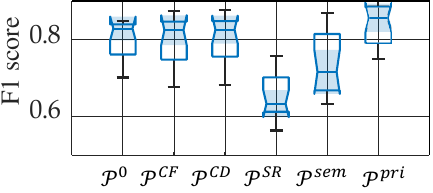}
		\end{minipage}
		\label{fig:limu_gru_f1}
	}
	\subfigure[UA: Accuracy]{
		\begin{minipage}[b]{0.22\textwidth}
			\includegraphics[width=1\linewidth]{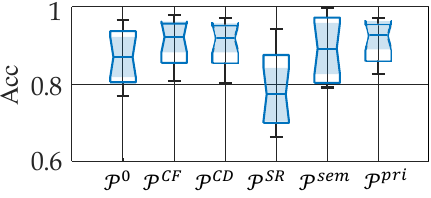}
		\end{minipage}
		\label{fig:limu_gru_acc_ua}
	}
	\subfigure[UA: F1 score]{
		\begin{minipage}[b]{0.22\textwidth}
			\includegraphics[width=1\linewidth]{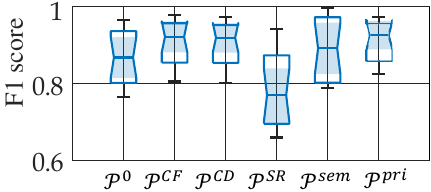}
		\end{minipage}
		\label{fig:limu_gru_f1_ua}
	}
	\caption{Boxplots of FUP performance based on LIMU-GRU, where Prism can outperform all other candidate methods. Note that LIMU-GRU is the SOTA base model for IMU data prediction. However, it does not perform best as the best model on Prism for the FUP problem.}
	\label{fig:limu_gru_fine}
\end{figure*}

\subsection{Experiments on Large-scale Non-\emph{i.i.d.} Dataset}
%\subsubsection{Large-scale Non-\emph{i.i.d.} Dataset based on SHL}

We further evaluate Prism on large-scale non-\emph{i.i.d.} IMU datasets. Specifically, we consider the University of Sussex-Huawei Locomotion (SHL) V1 dataset \cite{gjoreski2017versatile} a representative real-world IMU dataset. Four HUAWEI Mate 9 smartphones were respectively placed on four different body locations of a participant, including hand (ha), torso (to), backpack (ba), and trousers' front pocket (fr). 
% The data collection campaign employed three full-time participants. Four HUAWEI Mate 9 smartphones were respectively placed on four different body locations of a participant, including hand (ha), torso (to), backpack (ba), and trousers' front pocket (fr). 
A data logging application \cite{ciliberto2017high} was used to automatically log 16 sensor modalities including IMU sensors at a sampling rate of 100 Hz. During post-processing, an annotation tool is developed to help participants to label their activity as \emph{Car}, \emph{Bus}, \emph{Train}, \emph{Subway}, \emph{Walk}, \emph{Run}, \emph{Bike}, and \emph{Still}. 
%In addition, for some of the categories participants can choose the location (inside or outside) and the posture (stand or sit), which gives 18 combinations in total. As a result, the SHL V1 dataset contains three days of richly-annotated locomotion data.
We first pre-process the 9-dimension IMU data of three types of sensors, \emph{i.e.}, accelerometer, gyroscope, and magnetometer. Specifically, IMU data are segmented into samples of 5 seconds and then normalized as in Section \ref{sec:dataset}. We omit those samples with multiple inconsistent labels and obtain a dataset of 287,124 original IMU samples. In this study, we take a natural data partition scheme according to the phone location and derive four subsets, denoted as $\Gamma_{ha}$, $\Gamma_{to}$, $\Gamma_{ba}$, and $\Gamma_{fr}$, respectively. 
In addition, to mimic data sources with different sampling rates, we further equally divide $\Gamma_{ha}$ into four subsets, and downsample them with four sampling rates, \emph{i.e.}, 25 Hz, 50 Hz, 75 Hz, and 100 Hz, respectively. After that, we obtain four new subsets, denoted as $\Gamma_{ha}^{25}$, $\Gamma_{ha}^{50}$, $\Gamma_{ha}^{75}$, and $\Gamma_{ha}^{100}$. The same procedure repeats for each of the other three subsets, \emph{i.e.}, $\Gamma_{to}$, $\Gamma_{ba}$, and $\Gamma_{fr}$, too. As a result, we can obtain a manual data partition of 16 prior semantic domains, denoted as $\mathcal{P}^{\text{sem}}$, according to two types of metadata, \emph{i.e.}, the phone location and the sampling rate. We divide each domain into a training set, a validation set, and a testing set with a ratio of 6:2:2.

We conduct FUP evaluation on the testing sets of 16 domains in SHL. We considers 4 popular IMU base models, \emph{i.e.}, DCNN~\cite{yang2015deep}, GRU, LIMU-CNN, and LIMU-GRU~\cite{xu2021limu} in this experiment. 
% To ensure a fair evaluation, all classifiers and baseline models are trained with the same training hyper-parameters. Grid search is used to choose the hyper-parameters of the base models. 
Table~\ref{tab:overall_shl} shows the average overall performance of candidate methods on different base models with both the AR task and the UA task. 
%For more detailed results, Figure~\ref{fig:dcnn_fine} shows the box plots of FUP accuracy of different partition schemes based on DCNN model on 16 test domains.
%Table~\ref{tab:overall} shows the average overall performance of candidate methods on different base models with both the  task and the UA task.
We show the FUP performance based on DCNN~\cite{yang2015deep}, GRU, LIMU-CNN, and LIMU-GRU~\cite{xu2021limu} in Figure~\ref{fig:dcnn_fine}, Figure~\ref{fig:gru_fine}, Figure~\ref{fig:limu_cnn_fine} and Figure~\ref{fig:limu_gru_fine}, respectively. 
We can see that Prism outperforms all candidate methods on all base models on average. This is because Prism considers data partition based on the performance of training on the downstream task. 
% Such a partition may be beneficial to the downstream task or not, resulting in unstable end-to-end performance.
We further take the performance of evaluations based on DCNN on the AR task as an example, which is shown in Figure~\ref{fig:dcnn_acc} and Figure~\ref{fig:dcnn_f1}. We can see that the box plot of Prism is both higher and more compact, suggesting its superior performance across all test domains compared to other methods. Specifically, Prism can achieve an average increase of 9.5\% and 9.3\% on accuracy and F1 score compared with $\mathcal{P}^0$, respectively. The results demonstrate that Prism estimates the latent domains and the models can be trained well on corresponding domains. We can also see that all the partition-based methods except for $\mathcal{P}^{\text{SR}}$ (which lacks prior semantic information when testing) can outperform $\mathcal{P}^0$, which considers all samples as one domain. In fact, $\mathcal{P}^{\text{CF}}$ and $\mathcal{P}^{\text{sem}}$ can outperform $\mathcal{P}^{0}$ by 6.0\% and 3.8\% on F1 score, respectively. This is because data partition relieves the non-\emph{i.i.d.} issue in the dataset in some degree. 
% The performance of the prior-based unrealistic method $\mathcal{P}^{\text{sem}}$ is even not as well as the automatic methods, \emph{i.e.}, $\mathcal{P}^{\text{CF}}$ and $\mathcal{P}^{\text{CD}}$, in some cases. 
% This indicates that the prior semantic information differs from the true latent domains and may not always favor the inference task. 
%\subsubsection{FUP Accuracy on Different Test Domains}
We can also see difficulty differences in different test domains as there are always some test domains showing a low inference accuracy. For example, in Figure \ref{fig:dcnn_acc}, the typical $\mathcal{P}^0$ method can have a performance difference of over 17.5\%
%The range of the P^0 box in Figure 13(a) seems not larger than 20\%. The box plot of Prism appears not much shorter than others.%
between the easiest and the most difficult domain, \emph{i.e.}, the range between the upper and lower limits of the boxes. 

\subsection{System Costs}

\subsubsection{Training Costs}
 We investigate the system cost of different schemes from the following three aspects, \emph{i.e.}, total number of parameters, disk size, and memory consumption during inference. Table \ref{tab:real_world} demonstrates that the increase on system costs of Prism compared with $\mathcal{P}^0$ is acceptable. This is owing to the design of unified backbone $\mathcal{M}^{\text{enco}}$ in $\mathcal{M}_i$, indicating the feasibility of Prism's deployment.

 \begin{table}[]
 	\caption{Costs of all partition methods.}
 	\centering
 	\label{tab:real_world}
 	\scalebox{1}{
 		\begin{tabular}{ c c c c c c }
 			\toprule
 			Methods        & $\mathcal{P}^0$ & $\mathcal{P}^{\text{sem}}$ & $\mathcal{P}^{\text{CD}}$ & $\mathcal{P}^{\text{CF}}$ & $\mathcal{P}^{\text{pri}}$ \\ \hline
 			Parameters (KB)    & 96.7  & 386.8 & 290.1  & 360.2  & 118.5  \\  
 			Disk size (KB)     & 380 & 1520  & 1140  & 1490  & 468  \\
 			Memory (MB) & 119  & 714 & 833 & 894 & 152  \\ \bottomrule
 		\end{tabular}
 	}
 \end{table}
 
 \begin{figure}[]
	\centering
	\includegraphics[height=2.2cm]{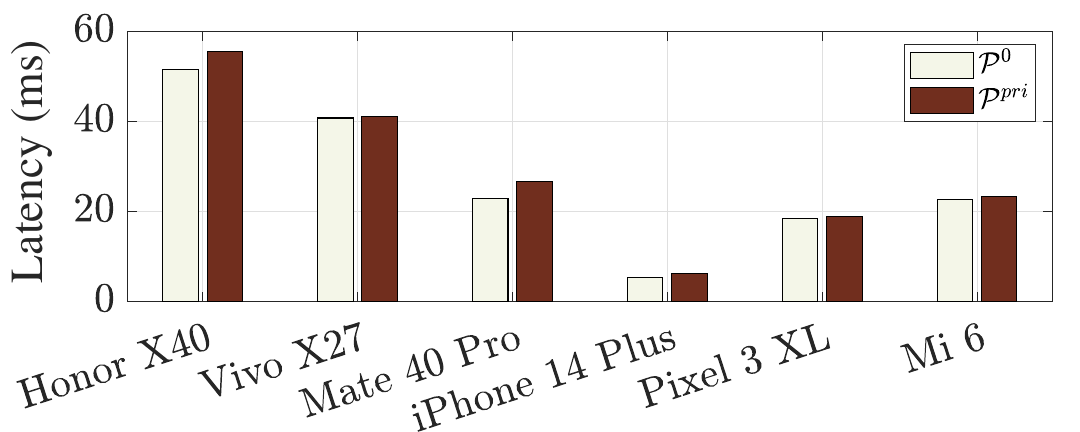}
	\caption{Inference latency of Prism and $\mathcal{P}^0$ on 6 typical mobile devices.}
	\label{fig:latency}
\end{figure}

\subsubsection{Inference Latency}
We evaluate the inference latency of Prism and the fast candidate method (\emph{i.e.}, $\mathcal{P}^0$) on different mobile devices. The results are shown in Figure \ref{fig:latency}. The results reveal that Prism only exhibits a negligibly higher latency when compared to the fastest scheme. This is because despite Prism utilizes multiple downstream task classifiers (\emph{e.g.}, $\mathcal{M}^{\text{task}}_i$ for $i \in [1,n]$), only one downstream task classifier (\emph{e.g.}, $\mathcal{M}^{\text{task}}_3$) needs to be executed during each inference. As a result, the latency of Prism is close to that of $\mathcal{P}^0$. We can also see that even on our lowest-end smartphones (\emph{i.e.}, Honor X40), the latency of Prism is below 60 ms. This highlights that Prism is lightweight and can be easily deployed to a wide variety of mobile devices with limited computational capacity.

\section{Discussion}

\textbf{Differences between Prism and MoE.} MoE, \emph{i.e.}, Mixture of Experts, is a popular technology to extend the model parameters and is similar with Prism. Prism differs from MoE in two aspects. First, experts in MoE are diversified just by constraints of losses, but they themselves cannot be related to the domains in the dataset. Second, models based on MoE architecture can only be deployed with the entire model, which is unacceptable for mobile applications. On the contrary, the models in Prism can be partially deployed \cite{li2024anole}, \emph{i.e.}, only models related to testing scenarios to be deployed, making Prism more lightweight and suitable for mobile devices.

\section{Related Work}

%\textbf{Machine Learning for IMU data.} Machine learning has been widely used for applications based on IMU data~\cite{zhu2017shakein, zhang2020smartso, shi2021face, xu2020touchpass}. 
%Traditional machine learning models are first used for IMU data inference, \emph{e.g.}, Hidden Markov Model (HMM)~\cite{xu2016air}, Support Vector Machine (SVM)~\cite{zhu2017shakein, zhang2020smartso} and Dynamic Time Warping (DTW)~\cite{xu2021novel}. These methods are easily deployed and work well with a small number of data samples. However, they all rely on manually designed features.
%For automatic feature extraction, deep-learning-based models are used for deep neural network designing for IMU data. CNN-based models~\cite{yang2015deep, ding2019deep} with a strong feature extraction capability are first used for inference on IMU data. RNN-based models~\cite{li2021cross, xu2021limu} are also useful with a strong generalization capability. Transformer-based models~\cite{xu2021limu} can also be used for IMU data inference with pre-training~\cite{tang2020exploring, Qian2022}. 

\subsection{Flexible User Perception for IMU Data.} 

Flexible user perception for IMU data has been widely explored with transfer-learning-based solutions~\cite{zhao2011cross, li2021cross}. However, these methods based on transfer learning do not consider the mobile setting, where the test domains are unknown. 
The domain partition is therefore proposed to solve the FUP problem~\cite{niu2020billion, li2021hermes, fang2019teamnet}. TeamNet~\cite{fang2019teamnet} explores and trains multiple small NNs through competitive and selective learning. UniHAR \cite{xu2023practically} adapts to all seen domains offline to ensure inference performance.
All of these methods require \emph{accurate} apriori information for data partition, which is hard to obtain in the real-world setting.

\subsection{Automatic Domain Estimation.} 

As for automatic domain estimation, a natural idea is to perform clustering before training, \emph{e.g.}, Clustered partition~\cite{5360282}. The key lies in the similarity measurement including mainly two types, \emph{i.e.}, prior information, and historical samples. First, through prior information, similarity graphs bring similar domains close to each other based on domain knowledge~\cite{kato08, han14}. However, such prior information constructed according to domain-based knowledge is also not easy to obtain, hindering the wide use of such approaches in real-world applications. 
Second, samples are used in data partition for more automatic clustering~\cite{zhang2016self, liu17}. 
However, these methods merely rely on the samples or features, which can result in missing intrinsic information, as reported in various real-world applications~\cite{zheng2019metadata, de18}. 
Third, the downstream task labels can also be used for the task-specific data partition~\cite{zheng2022towards, zheng2022life}. A task-oriented data grouping strategy based on the greedy method is proposed by TForest~\cite{zheng2022towards}. 
LEON~\cite{zheng2022life} proposed an online updating method for task-specific data partition. 
However, prior information is still needed for the initial data partition. DIVERSIFY \cite{lu2023out} iteratively estimates dynamic task-independent distributions of time series.
In contrast, Prism differs from the existing methods for it is automatic, prior-free, and task-aware.

\subsection{Quantification of Non-\emph{i.i.d.} Degree.} 

Non-\emph{i.i.d.} issue has been a research focus in the field of data mining for a long time \cite{wang2023distribution, chen2023enhancing, li2022federated}. The quantification of non-\emph{i.i.d.} degree between two distinct datasets can be computed as their difference of features of the same class between datasets \cite{he2021towards}. For the non-\emph{i.i.d.} index of one single dataset, prediction confidence is always considered as a flag for non-\emph{i.i.d.} testing \cite{bhardwaj2022ekya}. RISE \cite{zhai2021rise} proposed a non-\emph{i.i.d.} index based on Conformal Prediction (CP) theory \cite{shafer2008tutorial} for traditional machine learning models. However, RISE relies on the closed-form solution of the model and as a result unsuitable for deep neural network.
In contrast, Prism defines NID based on the difference of features in multiple partitions of the dataset, which is simple but effective.

\section{Conclusion}
In this paper, we have proposed a flexible user perception scheme, called Prism, for flexible user perception on mobile devices. Prism can automatically discover latent domains in a dataset with respect to a specific perception task, resulting in a set of domain-specific reliable task models for use.
As a result, Prism can obtain state-of-the-art prediction accuracy while having no particular requirements on how users operate their devices. Prism is lightweight and can be easily implemented on various mobile devices at a low cost. Extensive experiment results demonstrate that Prism can achieve the best flexible user perception performance at low latency.

\section*{Acknowledgments}
This work was supported in part by the Natural Science Foundation of China (Grants No. 62432008, 62472083), Natural Science Foundation of Shanghai (Grant No. 22ZR1400200) and Ant Group Research Fund (Grant No. 2021110892158).

% \newpage

\bibliographystyle{IEEEtran}
\bibliography{ref.bib}
\end{document}